%% file: main.tex
\documentclass[runningheads]{llncs}
\usepackage{graphicx}

\usepackage{tikz}
\usepackage{comment}
\usepackage{amsmath,amssymb} 
\usepackage{color}
\usepackage[misc]{ifsym}

\usepackage{booktabs}
\usepackage{multirow}
\newcommand{\keypoint}[1]{\noindent\textbf{#1}\quad}
\usepackage{booktabs}
\usepackage{floatrow}
\usepackage{float}
\floatstyle{plaintop}
\restylefloat{table}
\newfloatcommand{capbtabbox}{table}[][\FBwidth]
\usepackage[font=small,labelfont=bf,tableposition=top]{caption}
\usepackage{blindtext}

\usepackage{pifont}
\newcommand{\cmark}{\ding{51}}%
\newcommand{\xmark}{\ding{55}}%

\DeclareMathOperator*{\argmax}{arg\,max}

\makeatletter
\def\@fnsymbol#1{\ensuremath{\ifcase#1\or \textsuperscript{\Letter}\or \ddagger\or
   \mathsection\or \mathparagraph\or \|\or **\or \dagger\dagger
   \or \ddagger\ddagger \else\@ctrerr\fi}}
\makeatother

\makeatletter
\renewcommand\paragraph{
  \@startsection{paragraph} 
  {4} 
  {\z@} 
  {.5em \@plus1ex \@minus.2ex} 
  {-1.5em} 
  {\normalfont\normalsize\bfseries} 
}
\makeatother

\usepackage[pagebackref=false, breaklinks=true,letterpaper=true, colorlinks,citecolor=citecolor, linkcolor=linkcolor, bookmarks=false]{hyperref}
\definecolor{citecolor}{HTML}{0071BC}
\definecolor{linkcolor}{HTML}{ED1C24}
\newcommand{\subfig}[1]{\textcolor{linkcolor}{-#1}}

\usepackage[capitalize]{cleveref}
\crefname{section}{Sec.}{Secs.}
\Crefname{section}{Section}{Sections}
\Crefname{table}{Table}{Tables}
\crefname{table}{Tab.}{Tabs.}

\makeatletter
\DeclareRobustCommand\onedot{\futurelet\@let@token\@onedot}
\def\@onedot{\ifx\@let@token.\else.\null\fi\xspace}

\def\eg{\emph{e.g}\onedot} 
\def\ie{\emph{i.e}\onedot} 
\def\cf{\emph{cf}\onedot} 
\def\etc{\emph{etc}\onedot} \def\vs{\emph{vs}\onedot}

\makeatother

\begin{document}
\pagestyle{headings}
\mainmatter
\def\ECCVSubNumber{222}  
\title{Panoptic Scene Graph Generation}

%
\author{Jingkang Yang\inst{1} \and
Yi Zhe Ang\inst{1} \and
Zujin Guo\inst{1} \and\\
Kaiyang Zhou\inst{1} \and
Wayne Zhang\inst{2} \and
Ziwei Liu\inst{1}~\textsuperscript{\Letter}}
\authorrunning{J. Yang et al.}
%
\institute{S-Lab, Nanyang Technological University, Singapore\\
\email{\{jingkang001,yizhe.ang,gu00008,kaiyang.zhou,ziwei.liu\}@ntu.edu.sg}\and
SenseTime Research, Shenzhen, China\\
\email{wayne.zhang@sensetime.com}}
\maketitle
\input{sections/0_abstract}
\input{sections/1_introduction}

\input{sections/2_related_work}
\input{sections/3_problem_statement}

\input{sections/4_baseline_twostage}

\input{sections/4_baseline_onestage}
\input{sections/5_method_psgformer}
\input{sections/6_experiment}

\section*{Acknowledgements}
This work is supported by NTU NAP, MOE AcRF Tier 2 (T2EP20221-0033), and under the RIE2020 Industry Alignment Fund – Industry Collaboration Projects (IAF-ICP) Funding Initiative, as well as cash and in-kind contribution from the industry partner(s).

\newpage

%
%
\bibliographystyle{splncs04}
\bibliography{egbib}
\newpage
\input{sections/appendix}

\end{document}

%% file: sections/0_abstract.tex
\begin{abstract}
Existing research addresses scene graph generation (SGG)---a critical technology for scene understanding in images---from a detection perspective, \ie, objects are detected using bounding boxes followed by prediction of their pairwise relationships. We argue that such a paradigm causes several problems that impede the progress of the field. For instance, bounding box-based labels in current datasets usually contain redundant classes like hairs, and leave out background information that is crucial to the understanding of context. In this work, we introduce \emph{panoptic scene graph generation (PSG)}, a new problem task that requires the model to generate a more comprehensive scene graph representation based on panoptic segmentations rather than rigid bounding boxes. A high-quality \emph{PSG dataset}, which contains 49k well-annotated overlapping images from COCO and Visual Genome, is created for the community to keep track of its progress. For benchmarking, we build four two-stage baselines, which are modified from classic methods in SGG, and two one-stage baselines called PSGTR and PSGFormer, which are based on the efficient Transformer-based detector, \ie, DETR. 
While PSGTR uses a set of queries to directly learn triplets, PSGFormer separately models the objects and relations in the form of queries from two Transformer decoders, followed by a prompting-like relation-object matching mechanism. In the end, we share insights on open challenges and future directions. We invite users to explore the PSG dataset on our project page \url{https://psgdataset.org/}, and try our codebase \url{https://github.com/Jingkang50/OpenPSG}. 
\end{abstract}

%% file: sections/1_introduction.tex
\section{Introduction}
\label{sec:intro}

\input{figures/psg_vs_vg}

The goal of scene graph generation (SGG) task is to generate a graph-structured representation from a given image to abstract out objects---grounded by bounding boxes---and their pairwise relationships~\cite{sggsurvey21arxiv,sggsurvey20tnnls}. 
Scene graphs aim to facilitate the understanding of complex scenes in images and has potential for a wide range of downstream applications, such as image retrieval~\cite{johnson2015image,schuster2015generating,qi2017online}, visual reasoning~\cite{aditya2018image,shi2019explainable}, visual question answering~(VQA)~\cite{hildebrandt2020scene}, image captioning~\cite{gao2018image,chen2020say}, structured image generation and outpainting~\cite{johnson2018image,dhamo2020semantic,yang2022scene}, and robotics~\cite{gadre2022continuous,amiri2022reasoning}.

Since the introduction of SGG~\cite{johnson2015image}, this problem has been addressed from a detection perspective, \ie, using bounding boxes to detect objects followed by the prediction of their pairwise relationships~\cite{vg17ijcv,xu2017scene,liang2019vrr}. We argue that such a bounding box-based paradigm is not ideal for solving the problem, and would instead cause a number of issues that impede the progress of the field. Firstly, bounding boxes---as labeled in current datasets~\cite{vg17ijcv}---only provide a coarse localization of objects and often contain noisy/ambiguous pixels belonging to different objects or categories (see the bounding boxes of the two persons in Fig.~\ref{fig:psg_vs_vg}\subfig{-a}). Secondly, bounding boxes typically cannot cover the full scene of an image. For instance, the pavement region in Fig.~\ref{fig:psg_vs_vg}\subfig{-a} is crucial for understanding the context but is completely ignored. Thirdly, current SGG datasets often include redundant classes and information like \texttt{woman-has-hair} in Fig.~\ref{fig:psg_vs_vg}\subfig{-a}, which is mostly deemed trivial~\cite{liang2019vrr}. Furthermore, inconsistent and redundant labels are also observed in current datasets, \eg, the trees and benches in Fig.~\ref{fig:psg_vs_vg}\subfig{-a} are labeled multiple times, and some extra annotations do not contribute to the graph (see isolated nodes). Using such labels for learning might confuse the model.

Ideally, the grounding of objects should be clear and precise, and a scene graph should not only focus on salient regions and relationships in an image but also be comprehensive enough for scene understanding. We argue that as compared to bounding boxes, panoptic segmentation~\cite{panopticsegmentation} labels would be a better choice for constructing scene graphs. To this end, we introduce a new problem, \emph{panoptic scene graph generation}, or PSG, with a goal of generating scene graph representations based on panoptic segmentations rather than rigid bounding boxes.

To help the community keep track of the research progress, we create a new PSG dataset based on COCO~\cite{lin2014microsoft} and Visual Genome (VG)~\cite{vg17ijcv}, which contains 49k well-annotated images in total. We follow COCO's object annotation schema of 133 classes; comprising 80 thing classes and 53 stuff (background) classes. To construct predicates, we conduct a thorough investigation into existing VG-based datasets, \eg, VG-150~\cite{xu2017scene}, VrR-VG~\cite{liang2019vrr} and GQA~\cite{hudson2019gqa}, and summarize 56 predicate classes with minimum overlap and sufficient coverage of semantics. See Fig.~\ref{fig:psg_vs_vg}\subfig{-b} for an example of our dataset. From Fig.~\ref{fig:psg_vs_vg}, it is clear that the panoptic scene graph representation---including both panoptic segmentations and the scene graph---is much more informative and coherent than the previous scene graph representation.

For benchmarking, we build four two-stage models by integrating four classic SGG methods~\cite{xu2017scene,zellers2018neural,tang2018vctree,suhail2021energybased} into a classic panoptic segmentation framework~\cite{kirillov2019panoptic}. We also turn DETR~\cite{detr}, an efficient Transformer-based detector, into a one-stage PSG model dubbed as PSGTR, which has proved effective for the PSG task. We further provide another one-stage baseline called PSGFormer, which extends PSGTR with two improvements: 1) modeling objects and relations separately in the form of queries within two Transformer decoders, and 2) a prompting-like interaction mechanism. A comprehensive comparison of one-stage models and two-stage models is discussed in our experiments.

In summary, we make the following contributions to the SGG community:
\begin{itemize}
  \item \textbf{A New Problem and Dataset}: We discuss several issues with current SGG research, especially those associated with existing datasets. To address them, we introduce a new problem that combines SGG with panoptic segmentation, and create a large PSG dataset with high-quality annotations.
  \item \textbf{A Comprehensive Benchmark}: We build strong two-stage and one-stage PSG baselines, and evaluate them comprehensively on our new dataset, so that the PSG task is solidified in its inception.
  We find that one-stage models, despite having a simplified training paradigm, have great potential for PSG as it achieves competitive results on the dataset.
\end{itemize}

%% file: figures/psg_vs_vg.tex
\begin{figure}[t]
    \centering
    \includegraphics[width=0.7\linewidth]{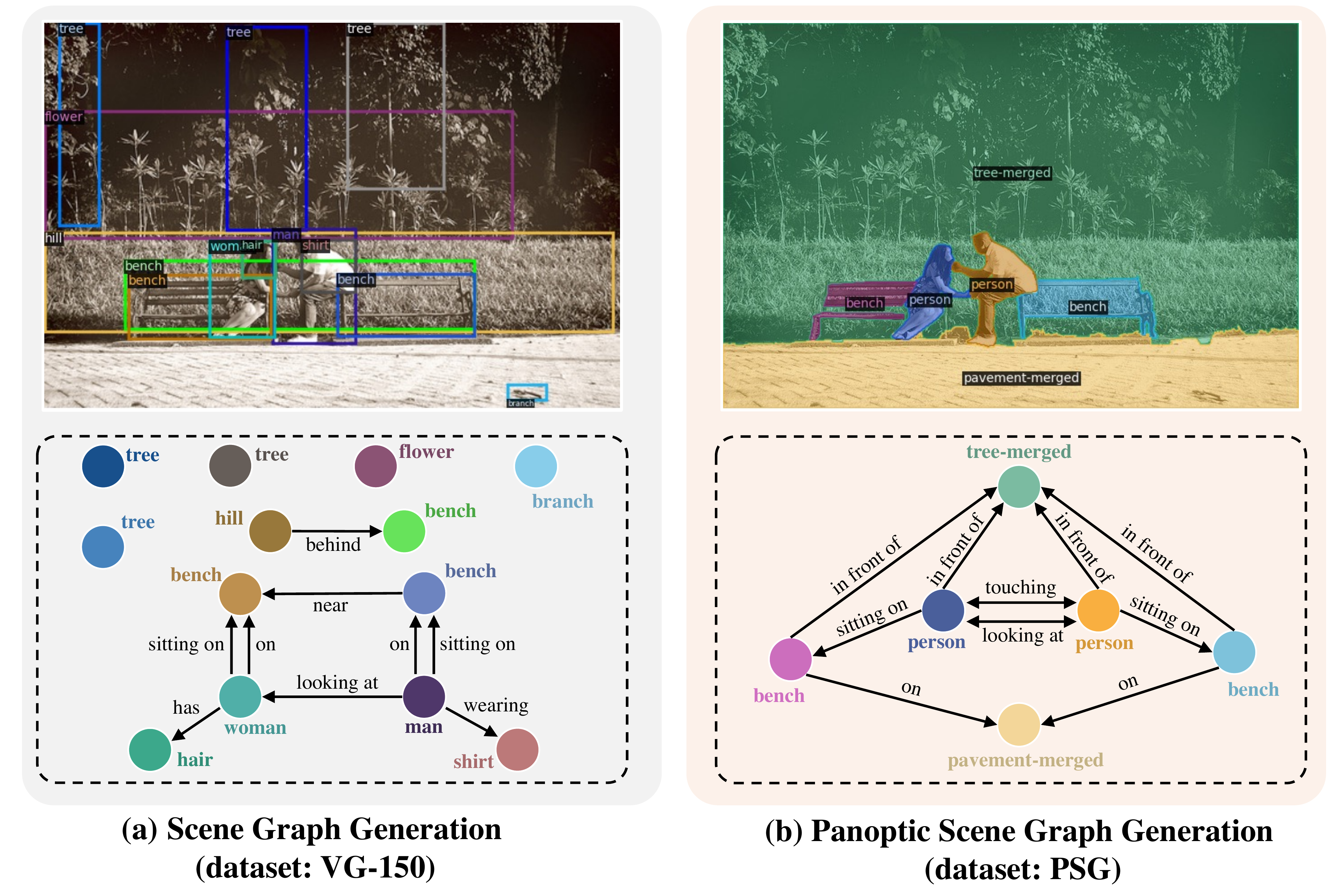}
    \caption{\textbf{Scene graph generation (a. SGG task) \vs panoptic scene graph generation (b. PSG task)}. The existing SGG task in (a) uses bounding box-based labels, which are often inaccurate---pixels covered by a bounding box do not necessarily belong to the annotated class---and cannot fully capture the background information. In contrast, the proposed PSG task in (b) presents a more comprehensive and clean scene graph representation, with more accurate localization of objects and including relationships with the background (known as stuff), \ie, the trees and pavement.
    }
    \label{fig:psg_vs_vg}
\end{figure}

%% file: sections/2_related_work.tex
\section{Related Work}
\label{sec:related_work}
\paragraph{Scene Graph Generation}
Existing scene graph generation (SGG) methods have been dominated by the two-stage pipeline that consists of object detection and pairwise predicate estimation. Given bounding boxes, early work predicts predicates using conditional random fields~\cite{johnson2015image,dai2017detecting} or casts predicate prediction into a classification problem~\cite{zhang2017relationship,kolesnikov2019detecting,qi2019attentive}. Inspired by knowledge graph embeddings, VTransE~\cite{zhang2017visual} and UVTransE~\cite{hung2020contextual} are proposed for explicit predicate modeling. Follow-up works have investigated various variants based on, \eg, RNN and graph-based modeling~\cite{xu2017scene,zellers2018neural,tang2018vctree,yang2018graph,lin2020gps,chen2019knowledge,li2018factorizable}, energy-based models~\cite{suhail2021energybased}, external knowledge~\cite{tang2018vctree,gu2019scene,zareian2020bridging,ZareianWYC20,lu2016visual}, and more recently language supervision~\cite{zhong2021learning,ye2021linguistic}. Recent research has shifted the attention to problems associated with the SGG datasets, such as the long-tailed distribution of predicates~\cite{tang2020unbiased,desai2021learning}, excessive visually-irrelevant predicates~\cite{liang2019vrr}, and inaccurate localization of bounding boxes~\cite{khandelwal2021segmentation}. In particular, a very recent study~\cite{khandelwal2021segmentation} shows that training an SGG model to simultaneously generate scene graphs and predict semantic segmentation masks can bring about improvements, which inspires our research. In our work, we study panoptic segmentation-based scene graph generation in a more systematic way by formulating a new problem and building a new benchmark.
We also notice that a closely-related topic human-object interaction (HOI)~\cite{gupta2015visual} shares a similar goal with SGG, \ie, to detect prominent relations from the image. However, the HOI task restricts the model to only detect human-related relations while ignoring the valuable information between objects that is often critical to comprehensive scene understanding. Nevertheless, many HOI methods are applicable to SGG tasks~\cite{gkioxari2018detecting,kato2018compositional,chao2018learning,wang2019deep,li2019transferable,zhou2020cascaded,wang2020learning,hou2020visual,li2020detailed,gao2020drg,kim2020uniondet,liu2020amplifying,tamura2021qpic,hou2021affordance,zhang2021mining,wang2022learning,zhang2022efficient}, and some of them have inspired our PSG baselines~\cite{kim2021hotr,zou2021end}.

\paragraph{Scene Graph Datasets}
\input{tables/psg}
While early SGG works have constructed several smaller-size datasets such as RW-SGD~\cite{johnson2015image}, VRD~\cite{lu2016visual}, and UnRel-D~\cite{peyre2017weakly}, the large-scale Visual Genome (VG)~\cite{vg17ijcv} quickly became the standard SGG dataset as soon as it was released in 2017, prompting subsequent work to research in a more realistic setting. 
However, several critical drawbacks of VG were raised by the community, and therefore, some VG variants were gradually introduced to address some problems.
Notice that VG contains an impractically large number of 33,877 object classes and 40,480 predicate classes, leading VG-150~\cite{xu2017scene} to only keep the most frequent 150 object classes and 50 predicate classes for a more realistic setting. 
Later, VrR-VG~\cite{liang2019vrr} argued that many predicates in VG-150 can be easily estimated by statistical priors, hence deciding to re-filter the original VG categories to only keep visually relevant predicate classes.
However, by scrutinizing VrR-VG, we find many predicates are redundant (\eg, \texttt{alongside}, \texttt{beside}, \texttt{besides}) and ambiguous (\eg, \texttt{beyond}, \texttt{down}).
Similar drawbacks appeared in another large-scale dataset with scene graph annotations called GQA~\cite{hudson2019gqa}.
In summary, although relations play a pivotal role in SGG tasks, a systematic definition of relations is unfortunately overlooked across all the existing SGG datasets.
Therefore, in our proposed PSG dataset, we consider both comprehensive coverage and non-overlap between words, and carefully define a predicate dictionary with 56 classes to better formulate the scene graph problem. Fig.~\ref{fig:predicates} shows the word cloud of the predicate classes, where font size indicates frequency.

Apart from the problem of predicate definition, another critical issue of SGG datasets is that they all adopt bounding box-based object grounding, which inevitably causes a number of issues such as coarse localization (bounding boxes cannot reach pixel-level accuracy), inability to ground comprehensively (bounding boxes cannot ground backgrounds), tendency to provide trivial information (current datasets usually capture objects like \texttt{head} to form the trivial relation of \texttt{person-has-head}), and duplicate groundings (the same object could be grounded by multiple separate bounding boxes). These issues together have caused the low-quality of current SGG datasets, which impede the progress of the field.
Therefore, the proposed PSG dataset tries to address all the above problems by grounding the images using accurate and comprehensive panoptic segmentations with COCO's appropriate granularity of object categories.
Table~\ref{tab:dataset_comparison} compares the statistics of the PSG dataset with classic SGG datasets.

\paragraph{Panoptic Segmentation}
The panoptic segmentation task unifies semantic segmentation and instance segmentation~\cite{panopticsegmentation} for comprehensive scene understanding, and the first approach is a simple combination of a semantic segmentation model and an instance segmentation model to produce stuff masks and thing masks respectively~\cite{panopticsegmentation}. Follow-up work, such as Panoptic FPN~\cite{kirillov2019panoptic} and UPSNet~\cite{Xiong2019UPSNetAU}, aim to unify the two tasks in a single model through multi-task learning to achieve gains in compute efficiency and segmentation performance. Recent approaches (e.g., MaskFormer~\cite{Cheng2021MaskFormer}, Panoptic Segformer~\cite{Li2021PanopticS} and K-Net~\cite{zhang2021knet}) have turned to more efficient architectures based on Transformers like DETR~\cite{detr}, which simplifies the detection pipeline by casting the detection task as a set prediction problem.

%% file: tables/psg.tex
\begin{figure*}[!t]\RawFloats
\captionsetup[table]{position=top}
\begin{minipage}{0.7\textwidth}
\captionof{table}{\textbf{Comparsion between classic SGG datasets and PSG dataset.} \#PPI counts predicates per image. DupFree checks whether duplicated object groundings are cleaned up. Spvn indicates whether the objects are grounded by bounding boxes or segmentations.
}
\label{tab:dataset_comparison}
\centering
\setlength\tabcolsep{2.5pt}

\resizebox{\textwidth}{!}{
\begin{tabular}{@{\hskip 0.05in}l@{\hskip 0.05in}|ccccccc@{\hskip 0.05in}}
\toprule
Dataset &\#Image &\#ObjCls &\#RelCls  & \#PPI & DupFree & Spvn & Source \\ \midrule
VG~\cite{vg17ijcv}         & 108K &  34K   &  40K     &  21.4   & \xmark  & BBox  & COCO \& Flickr\\ 
VG-150~\cite{xu2017scene}  & 88K  &  150   &   50      &  5.7   & \xmark  & BBox  & Clean VG \\ 
VrR-VG~\cite{liang2019vrr} & 59K  &  1,600 &  117      &  3.4   & \xmark  & BBox  & Clean VG \\ 
GQA~\cite{hudson2019gqa}   & 85K  &  1,703 &  310      &  50.6  & \cmark   & BBox & Re-annotate VG \\ 
\midrule
\textbf{PSG}               & 49K  & 133 &  56     &  5.7    & \cmark & Seg  & Annotate COCO\\ \bottomrule
\end{tabular}}
\end{minipage}
\hspace{6mm}
\begin{minipage}{0.23\textwidth}
\centering
\includegraphics[width=\linewidth]{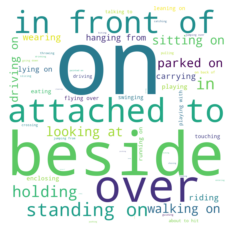}

\caption{Word Cloud for PSG Predicates.}
\label{fig:predicates}
\end{minipage}
\end{figure*}

%% file: sections/3_problem_statement.tex
\section{Problem and Dataset}
\label{sec:problem_dataset}

\keypoint{Recap: Scene Graph Generation}
We first briefly review the goal of the classic scene graph generation (SGG) task, which aims to model the distribution:
\begin{equation}
	\label{E:sgg_def}
	\operatorname{Pr}\left(\mathbf{G} \mid \mathbf{I}\right)=\operatorname{Pr}\left(\mathbf{B},\mathbf{O},\mathbf{R} \mid \mathbf{I}\right),
\end{equation}
where $\mathbf{I} \in \mathbb{R}^{H \times W \times 3}$ is the input image, and $\mathbf{G}$ is the desired scene graph which comprises the bounding boxes $\mathbf{B}=\left\{\mathbf{b}_{1}, \ldots, \mathbf{b}_{n}\right\}$ and labels $\mathbf{O}=\left\{o_{1}, \ldots, o_{n}\right\}$ that correspond to each of the $n$ objects in the image, and their relations in the set $\mathbf{R}=\left\{r_{1}, \ldots, r_{l}\right\}$. 
More specifically, $\mathbf{b}_{i} \in \mathbb{R}^{4}$ represents the box coordinates, $o_{i} \in \mathbb{C}^\text{O}$ 
and $r_{i} \in \mathbb{C}^\text{R}$ 
belong to the set of all object and relation classes.

\medskip

\keypoint{Panoptic Scene Graph Generation} Instead of localizing each object by its bounding box coordinates, the new task of panoptic scene graph generation (PSG task) grounds each object with the more fine-grained panoptic segmentation. For conciseness, we refer to both objects and background as objects.

Formally, with panoptic segmentation, an image is segmented into a set of masks $\mathbf{M}=\left\{\mathbf{m}_{1}, \ldots, \mathbf{m}_{n}\right\}$, where $\mathbf{m}_{i} \in\{0,1\}^{H \times W}$. Each mask is associated with an object with class label $o_{i} \in \mathcal{C}^O$. A set of relations $\mathbf{R}$ between objects are also predicted. The masks do not overlap, \ie, $\sum_{i=1}^{n} \mathbf{m}_{i} \leq \mathbf{1}^{H \times W}$. Hence, PSG task models the following distribution:

\begin{equation}
	\label{E:psgg_def}
	\operatorname{Pr}\left(\mathbf{G} \mid \mathbf{I}\right)=\operatorname{Pr}\left(\mathbf{M},\mathbf{O},\mathbf{R} \mid \mathbf{I}\right).
\end{equation}


\paragraph{PSG Dataset}
To address the PSG task, we build our PSG dataset following these three major steps. Readers can find more details in the Appendix.

\noindent\textbf{Step 1: A Coarse COCO \& VG Fusion:} 
To create a dataset with both panoptic segmentation and relation annotations, we use the 48,749 images in the intersection of the COCO and VG datasets with an automatic but coarse dataset fusion process.
Specifically, we use an object category matching procedure to match COCO's segmentations with VG's bounding boxes, so that part of VG's predicates are applied to COCO's segmentation pairs.
Due to the inherent mismatch between the label systems and localization annotations of VG and COCO, the auto-generated dataset is very noisy and requires further cleaning.

\noindent\textbf{Step 2: A Concise Predicate Dictionary:}
Inspired by the appropriate granularity of COCO categories~\cite{lin2014microsoft}, we carefully identify 56 salient relations by taking reference from common predicates in the initial noisy PSG dataset and all VG-based datasets including VG-150~\cite{xu2017scene}, VrR-VG~\cite{liang2019vrr} and GQA~\cite{hudson2019gqa}.
The selected 56 predicates are maximally independent (\eg, we only keep ``\texttt{over/on}'' and do not have ``\texttt{under}'') and cover most common cases in the dataset.

\noindent\textbf{Step 3: A Rigorous Annotation Process:}
Building upon the noisy PSG dataset, we require the annotators to
\emph{1) filter} out incorrect triplets, and
\emph{2) supplement} more relations between not only object-object, but also object-background and background-background pairs, using the predefined 56 predicates.
To prevent ambiguity between predicates, we ask the annotators strictly not to annotate using general relations like \texttt{on}, \texttt{in} when a more precise predicate like \texttt{parked on} is applicable.
With this rule, the PSG dataset allows the model to understand the scene more precisely and saliently.

\noindent\textbf{Quality Control:} The PSG dataset goes through a professional dataset construction process. The main authors first annotated 1000 images to construct a detailed documentation (available in project page), and then employed a professional annotation company (sponsored by SenseTime) to annotate the training set within a month (US\$11K spent). Each image is annotated by two workers and examined by one head worker. All the test images are annotated by the authors.

\noindent\textbf{Summary:} Several merits worth highlighting by virtue of the novel and effective procedure of PSG dataset creation:
\emph{1) Good grounding annotation} from the pixel-wise panoptic segmentation from COCO dataset~\cite{lin2014microsoft},
\emph{2) Clear category system} with 133 objects (\ie, things plus stuff) and 56 predicates with appropriate granularity and minimal overlaps, and
\emph{3) Accurate and comprehensive relation annotations} from a rigorous  annotation process that pays special attention to salient relations between object-object, object-background and background-background.
These merits address the notorious shortcomings~\cite{li2022devil} of classic scene graph datasets discussed in Sec.~\ref{sec:related_work}.

\paragraph{Evaluation and Metrics}
This section introduces the evaluation protocol for the PSG task.
Following the settings of the classic SGG task~\cite{sggsurvey20tnnls,sggsurvey21arxiv}, our PSG task comprises two sub-tasks: \emph{predicate classification} (when applicable) and \emph{scene graph generation} (main task) to evaluate the PSG models.
\emph{Predicate classification (PredCls)} aims to generate a scene graph given the ground-truth object labels and localization. The goal is to study the relation prediction performance without the interference of the segmentation performance. Notice that this metric is only applicable to two-stage PSG models in Sec.~\ref{sec:two-stage-baseline}, since one-stage models cannot leverage the given segmentations to predict scene graph.
\emph{Scene graph generation (SGDet)} aims to generate scene graphs from scratch, which is the main result for the PSG task.

We also notice that classic SGG tasks contain another sub-task of scene graph classification (SGCls), which provide the ground-truth object groundings to simplify the scene graph generation process.
We find SGCls is not applicable for PSG baselines.
Unlike SGG tasks where object detectors such as Faster-RCNN~\cite{ren2015faster} can utilize ground-truth object bounding boxes to replace predictions from the Region Proposal Network (RPN), panoptic segmentation models are unable to directly use the ground-truth segmentations for classification, so the SGCls task is inapplicable even for two-stage PSG methods.

The classic metrics of \emph{R@K} and \emph{mR@K} are used to evaluate the previous two sub-tasks, which calculates the triplet recall and mean recall for every predicate category, given the top K triplets from the PSG model. Notice that PSG grounds objects with segmentation, a successful recall requires both subject and object to have mask-based IOU larger than 0.5 compared to their ground-truth counterparts, with the correct classification on every position in the S-V-O triplet.

While the triplet recall rates that mentioned above are the main metric for PSG task, since objects are required to be grounded by segmentation masks, panoptic segmentation metrics~\cite{panopticsegmentation} such as PQ~\cite{kirillov2019panoptic} can be used for model diagnosis. However, it is not considered as the core evaluation metric of PSG task.

%% file: sections/4_baseline_twostage.tex
\section{PSG Baselines}
To build a comprehensive PSG benchmark, we refer to frameworks employed in the classic SGG task~\cite{sggsurvey21arxiv,sggsurvey20tnnls} and prepare two-stage and one-stage baselines.

\subsection{Two-Stage PSG Baselines}
\label{sec:two-stage-baseline}
\input{figures/two_stage}
Most prior SGG approaches tackle the problem in two stages: first performing object detection using off-the-shelf detectors like Faster-RCNN~\cite{ren2015faster}, then pairwise relationship prediction between these predicted objects.
As shown in Figure~\ref{fig:two_stage}, we follow a similar approach in establishing two-stage baselines for the PSG task: 
\textbf{1)} using pretrained panoptic segmentation models of classic Panoptic FPN~\cite{kirillov2019panoptic} to extract initial object features, masks and class predictions, and then 
\textbf{2)} processing them using a relation prediction module from classic scene graph generation methods like IMP~\cite{xu2017scene}, MOTIFS~\cite{zellers2018neural}, VCTree~\cite{tang2018vctree}, and GPSNet~\cite{lin2020gps} to obtain the final scene graph predictions.
In this way, classic SGG methods can be adapted to solve the PSG task with minimal modification.
Formally, the two-stage PSGG baselines decompose formulation from Eq.~\ref{E:sgg_def} to Eq.~\ref{E:sgg_twostage_def}.

\begin{equation}
	\label{E:sgg_twostage_def}
	\operatorname{Pr}\left(\mathbf{G} \mid \mathbf{I}\right)=\operatorname{Pr}\left(\mathbf{M} \mid \mathbf{I}\right) \cdot \operatorname{Pr}\left(\mathbf{O} \mid \mathbf{M}, \mathbf{I}\right) \cdot \operatorname{Pr}\left(\mathbf{R} \mid \mathbf{O}, \mathbf{M}, \mathbf{I}\right).
\end{equation}


%% file: figures/two_stage.tex
\begin{figure}[t]
    \centering
    \includegraphics[width=\linewidth]{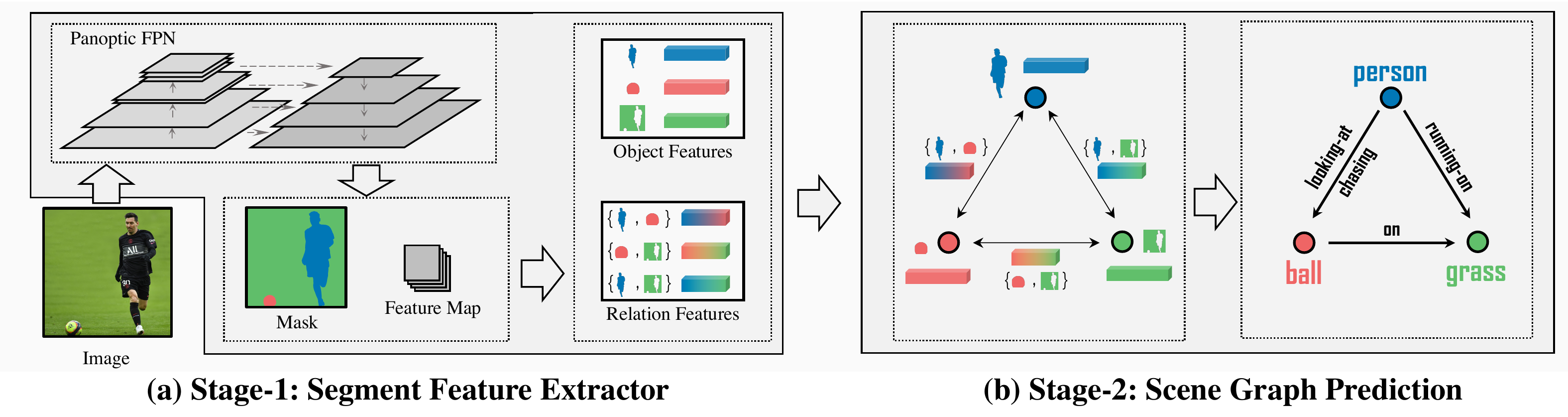}
    \caption{\textbf{Two-stage PSG baselines using Panoptic FPN.} \textbf{a)} In stage one, for each thing/stuff object, Panoptic FPN~\cite{kirillov2019panoptic} produces a segmentation mask with its tightest bounding box to crop out the object feature. The union of relevant objects can produce relation features. \textbf{b)} In the second stage, the extracted object and relation features are fed into by any existing SGG relation model to predict the relation triplets.}
    \label{fig:two_stage}
\end{figure}

%% file: sections/4_baseline_onestage.tex
\subsection{A One-Stage PSG Baseline - PSGTR}
\label{sec:one-stage-baseline}
Unlike classic dense prediction models (\eg, Faster-RCNN~\cite{ren2015faster}) with sophisticated design, the transformer-based architectures support flexible input and output specifications.
Based on the end-to-end DETR~\cite{detr} and its extension to the HOI task~\cite{zou2021end}, we naturally design a one-stage PSG method named PSGTR to predict triples and localizations simultaneously, which can be directly modeled as Eq.~\ref{E:psgg_def} without decomposition.

\paragraph{Triplet Query Learning Block}
As shown in Fig.~\ref{fig:one_stage}, PSGTR first extracts image features from a CNN backbone and then feeds the features along with queries and position encoding into a transformer encoder-decoder. Here we expect the queries to learn the representation of scene graph triplets, so that for each triplet query, the subject/predicate/object predictions can be extracted by three individual Feed Forward Networks (FFNs), and the segmentation task can be completed by two panoptic heads for subject and object, respectively.

\paragraph{PSG Prediction Block}
To train the model, we extend the DETR's Hungarian matcher~\cite{kuhn1955hungarian} into a triplet Hungarian matcher.
To match the triplet query $\mathcal{T}_i\in\mathbb{Q}^\text{T}$ with ground truth triplets $\mathcal{G}$, all contents in the triplet (\ie, all outputs that are predicted from $\mathcal{T}_i$) are used, including the class of subject $\ddot{\mathcal{T}}_i^\text{S}$, relation $\ddot{\mathcal{T}}_i^\text{R}$, and object $\ddot{\mathcal{T}}_i^\text{O}$, and localization of subjects $\tilde{\mathcal{T}}_i^\text{S}$ and objects $\tilde{\mathcal{T}}_i^\text{O}$.
Therefore, the triplet matching (tm) cost $\mathbf{C}_\text{tm}$ is designed with the combination of class matching $\mathbf{C}_\text{cls}$ and segments matching $\mathbf{C}_\text{seg}$:
%
\begin{equation}
\mathbf{C}_\text{tm}\left(\mathcal{T}_i,\mathcal{G}_{\sigma(i)}\right) = 
\sum_{k \in \{\text{S}, \text{O}\}} 
\mathbf{C}_\text{seg}\left(\tilde{\mathcal{T}}_i^k, \tilde{\mathcal{G}}_{\sigma(i)}^k\right) + 
\sum_{k \in \{\text{S}, \text{R}, \text{O}\}} \mathbf{C}_{\text{cls}}\left(\ddot{\mathcal{T}}_i^k, \ddot{\mathcal{G}}_{\sigma(i)}^k\right),
\end{equation}
where $\sigma$ is the mapping function to correspond each triplet query $\mathcal{T}_i\in\mathbb{Q}^\text{T}$ to the closest ground truth triplet. The triplet query set $\mathbb{Q}^\text{T}$ collects all the $|\mathbb{Q}^\text{T}|$ triplet queries. 
The optimization objective is thus:
\begin{equation}
\hat{\sigma} = \argmax_\sigma\sum_{i=1}^{|\mathbb{Q}^\text{T}|}\mathbf{C}_\text{tm}\left(\mathcal{T}_i,\mathcal{G}_{\sigma(i)}\right).
\end{equation}
Once the matching is done, the total loss $\mathcal{L}_\text{total}$ can be calculated by applying cross-entropy loss $\mathcal{L}_\text{cls}$ for labels and DICE/F-1 loss~\cite{milletari2016v} for segmentation $\mathcal{L}_\text{cls}$:
\begin{equation}
\label{E:total_loss}
\mathcal{L}_\text{total} = 
\sum_{i=1}^{|\mathbb{Q}^\text{T}|}
\left(\sum_{k \in \{\text{S}, \text{O}\}} 
\mathcal{L}_\text{seg}\left(\tilde{\mathcal{T}}_i^k, \tilde{\mathcal{G}}_{\hat{\sigma}(i)}^k\right)
+
\sum_{k \in \{\text{S}, \text{R}, \text{O}\}} \mathcal{L}_\text{cls}\left(\ddot{\mathcal{T}}_i^k, \ddot{\mathcal{G}}_{\hat{\sigma}(i)}^k\right)\right).
\end{equation}
\input{figures/one_stage}

%% file: figures/one_stage.tex
\begin{figure}[t]
    \centering
    \includegraphics[width=\linewidth]{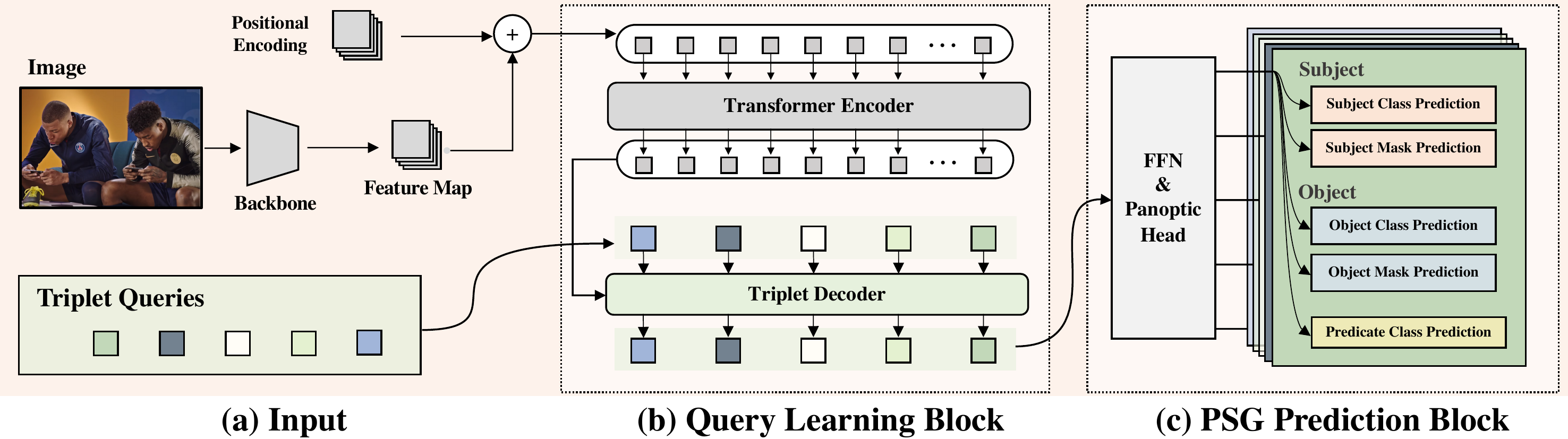}
    \caption{\textbf{PSGTR: One-stage PSG baseline.} The one-stage model takes in \textbf{a)} features extracted by CNNs with positional encoding, and a set of queries aiming to represent triplets. \textbf{b)} Query learning block processes image features with Transformer encoder-decoder and use queries to represent triplet information. Then, \textbf{c)} the PSG prediction head concretes the triplet predictions by producing subject/object/predicate classes using simple FFNs, and uses panoptic heads for panoptic segmentation.}
    \label{fig:one_stage}
\end{figure}

%% file: sections/5_method_psgformer.tex
\subsection{Alternative One-Stage PSG Baseline - PSGFormer}
\label{sec:psgformer}
\input{figures/psgformer}
Based on the PSGTR baseline that explained in Section~\ref{sec:one-stage-baseline}, we extend another end-to-end HOI method~\cite{kim2021hotr} and further propose the alternative one-stage PSG baseline named PSGFormer, featured by an explicit relation modeling with a prompting-like matching mechanism. The model diagram is illustrated in Figure~\ref{fig:psgformer} and will be elaborated as follows.

\paragraph{Object \& Relation Query Learning Block}
Compared to the classic object-oriented tasks such as object detection and segmentation, the most significant uniqueness of PSG task as well as SGG task is their extra requirements on the predictions of relations.
Notice that the relation modeling in our two-stage baselines depends on features from object-pairs, while PSGTR implicitly models the objects and relations altogether within the triplets, the important relation modeling has not been given a serious treatment.
Therefore, in the exploration of PSGFormer, we explicitly model the relation query $\mathcal{R}_i\in\mathbb{Q}^\text{R}$, as well as object query $\mathcal{O}_i\in\mathbb{Q}^\text{O}$ separately, in hope that object queries to specially pay attentions to objects (\eg, \texttt{person} and \texttt{phones}), and relation queries to focus on the area where the relationship takes place in the picture (\eg, \texttt{person} \texttt{looking-at} \texttt{phone}). Similar to PSGTR in Figure~\ref{fig:one_stage}, both object and relation queries with CNN features and position encoding are fed into a transformer encoder, but being decoded with their corresponding decoder, \ie, object or relation decoder, so that the queries can learn the corresponding representations.

\paragraph{Object \& Relation Query Matching Block}
In PSGFormer, each object query yields an object prediction with FFN and a mask prediction with a panoptic head, and each relation query yields a relation prediction. However, due to the parallel process of object queries and relation queries, the missing interdependence between different query types makes the triplet still not formed.
To connect object and relation queries for compositing triplets, we are inspired by the design in HOTR~\cite{kim2021hotr} and implement a prompting-like query matching block.

Query matching block models the triplet composition task as a fill-in-the-blank question with prompts, \ie, by prompting a relation, we expect a pair of suitable objects provided by their corresponding object queries can be selected, so that a complete \texttt{subject-predicate-object} triplet, can be generated.
Therefore, two selectors, \ie, subject selector and object selector, are required.

Given a relation query $\mathcal{R}_i\in\mathbb{Q}^{R}$ as prompt, both subject selector and object selector should return the most suitable candidate to form a complete triplet.
We use the most standard cosine similarity between object queries and the provided relation query and pick the highest similarity to determine the subject and object candidates.
It should also be noticed that subject and object selectors should rely on the level of association between objects and relation queries rather than the semantic similarity.
Besides, object queries are regarded as different roles (\ie, subject or object) in different selectors. 
Therefore, the object queries are expected to pass another two FFNs to extract some specific information for subject (with FFN denoted as $f_{\text{S}}$) and object (with FFN denoted as $f_{\text{O}}$), so that the distinguishable subject and object representations are obtained from the object queries.
With the idea above, a set of subjects $\mathbb{S}$ are generated in Eq.~\ref{E:sub_select}, with the $i$-th subject corresponding to the $i$-th relation query $\mathcal{R}_i$. With a similar process, the object set $\mathbb{O}$ is also generated.
\begin{equation}
\label{E:sub_select}
\mathbb{S} = \left\{\mathcal{S}_i~|~\mathcal{S}_i=\arg\max_{\mathcal{O}}\left(f_{\text{S}}(\mathcal{O}_j)\cdot\mathcal{R}_i\right),
~\mathcal{O}_j\in\mathbb{Q}^\text{O},
~\mathcal{R}_i\in\mathbb{Q}^\text{R}~\right\}.
\end{equation}

Till now, the subject set $\mathbb{S}$ and the object set $\mathbb{O}$ are well-prepared by subject and object selectors, with the $i$-th subject query $\mathcal{S}_i$ and the $i$-th object $\mathcal{O}_i$ corresponding to the $i$-th relation query $\mathcal{R}_i$.
Finally, it is straightforward to obtain all the matched triplet $\mathbb{T}$, which is shown in Eq.~\ref{E:t_i}.

\begin{equation}
\label{E:t_i}
\mathbb{T} = \left\{(\mathcal{S}_i, \mathcal{R}_i, \mathcal{O}_i)~|
~\mathcal{S}_i\in\mathbb{S},
~\mathcal{R}_i\in\mathbb{R},
~\mathcal{O}_i\in\mathbb{O}
\right\}.
\end{equation}

Apart from interpreting the query matching as a prompt-like process, it can also be considered as a cross-attention mechanism. 
For a relation query (Q), the goal is to find the high-attention relations among all subject / object representations, which are considered as keys (K). 
The subject / object labels predicted by the corresponding representations are regarded as values (V), so that the QKV attention model outputs the labels of selected keys. Fig.~\ref{fig:psgformer}\subfig{-c} is generally depicted following this interpretation.

\paragraph{PSG Prediction Block}
Similar to PSGTR, with the predicted triplets prepared, the prediction block can finally train the model using $\mathcal{L}_\text{total}$ from Eq.~\ref{E:total_loss}. In addition, with object labels and masks predicted by object queries, a standard training loss introduced in panoptic segmentation DETR~\cite{detr} is used to enhance the object decoder and avoid duplicate object groundings.

%% file: figures/psgformer.tex
\begin{figure}[t]
    \centering
    \includegraphics[width=\linewidth]{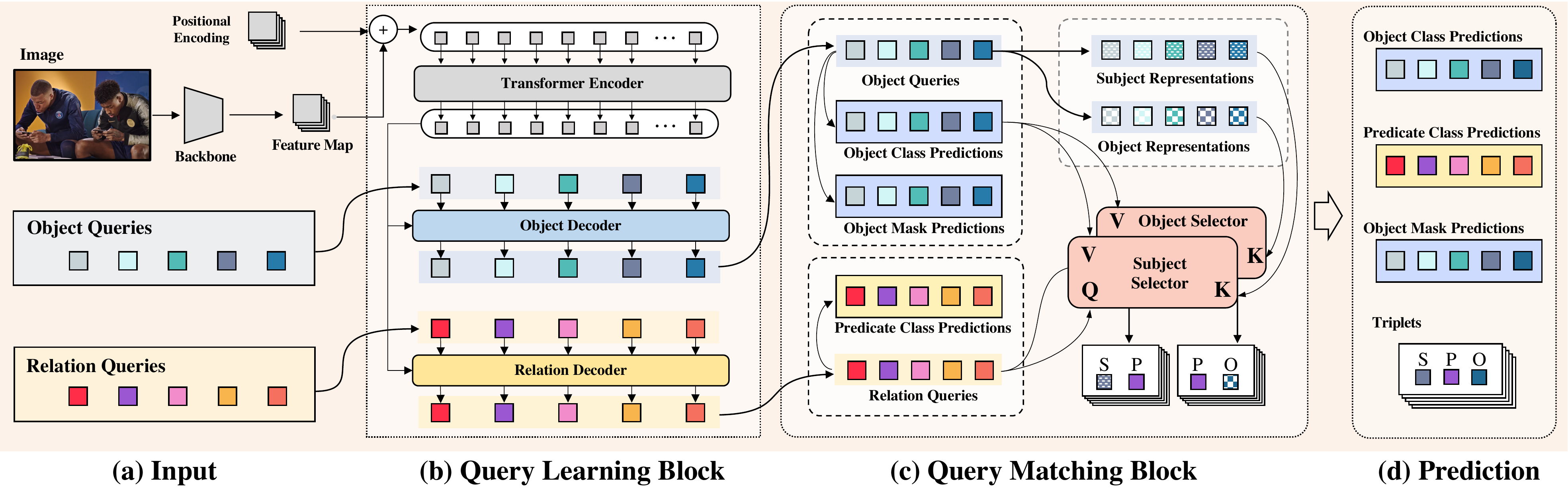}
    \caption{\textbf{PSGFormer: The proposed one-stage PSG method.} 
    \textbf{a)} Two types of queries, \ie, object queries and relation queries, are fed into transformer block with CNN features and positional encoding.
    \textbf{b)} Query Learning Block processes image features with one encoder and output object or relation queries with the corresponding decoder.
    \textbf{c)} Object queries and relation queries interact with each other in the prompting-like query matching block, so that the triplets are formed and proceed to 
    \textbf{d)} PSG prediction block to output results or compute loss as PSGTR behaves.}
    \label{fig:psgformer}
\end{figure}

%% file: sections/6_experiment.tex
\section{Experiments}
In this section, we first report the results of all PSG methods introduced in the paper. Implementation details are available in the appendix, and all codes are integrated in the \href{https://github.com/Jingkang50/OpenPSG}{OpenPSG} codebase, which is developed based on MMDetection~\cite{chen2019mmdetection}. Most of the two-stage SGG implementations refer to MMSceneGraph~\cite{wong2020mmsg} and Scene-Graph-Benchmark.pytorch~\cite{tang2020sggcode}.

\input{tables/exp_cls}
\subsection{Main Results}
Table~\ref{tab:exp_results} reports the scene graph generation performance of all the methods mentioned in Sec.~\ref{sec:two-stage-baseline}, Sec.~\ref{sec:one-stage-baseline}, and Sec.~\ref{sec:psgformer} under the PSG dataset.
Fig.~\ref{fig:exp_seg} reports the panoptic segmentation result using PQ and visualizes the segmentation results of two examples as well as the predicted scene graph in the form of triplet lists.

\paragraph{Two-Stage Baselines Rely on First-Stage Performance}
For predicate classification task (PredCls) that is only applicable to two-stage models, the provided ground-truth segmentation can significantly improve the triplet prediction performance.
For example, even the most classic method IMP can reach over 30\% R@20, which already exceeds all the available R@20 under the scene graph generation (SGDet) task (\cf 28.4\% by PSGTR).
This phenomenon indicates that a good panoptic segmentation performance could naturally benefit the PSG task. Further evidence where the performance of IMP on the SGDet task is almost halved (from 32\% to 17\% on R@20) strengthens the above conjecture.

\paragraph{Some SGG Techniques for VG are not Effective for PSG}
Table~\ref{tab:exp_results} shows that the results of some two-stage baselines (\ie, IMP, MOTIFS, and VCTree) are generally proportional to their performance on SGG tasks, indicating that the advantages of the two-stage models (\ie, MOTIFS and VCTree) are transferable to PSG task.
However, we notice that another two-stage baseline, GPSNet, does not seem to exceed its SGG baselines of MOTIFS and VCTree in the PSG task.
The key advantage of GPSNet over MOTIFS and VCTree is that it explicitly models the direction of predicates. While the design can be effective in the VG dataset where many predicates are trivial with obvious direction of predicates (\eg, \texttt{of} in \texttt{hair-of-man}, \texttt{has} in \texttt{man-has-head}), PSG dataset gets rid of these predicates, so the model may not be effective as expected. 

\input{figures/exp_seg}
\paragraph{PSGFormer is an Unbiased PSG Model}
When the training schedule is limited to 12 epochs, the end-to-end baseline PSGFormer outperforms the best two-stage model VCTree by significant 4.8\% on mR@20 and 8.5\% on mR@100.
Although PSGFormer still cannot exceed two-stage methods on the metrics of R@20/50/100, its huge advantage in mean recall indicates that the model is unbiased in predicting relations.
As Fig.~\ref{fig:exp_seg} shows, PSGFormer can predict unusual but accurate relations such as \texttt{person-going down-snow} in the upper example, and the imperceptible relation \texttt{person-driving-bus} in the lower example.
Also, in the upper example, PSGFormer predicts an interesting and exclusive triplet \texttt{person-wearing-skis}. This unique prediction should come from the design of the separate object / relation decoders, so that relation queries can independently capture the meaning of the predicates.

\paragraph{PSGTR Obtains SOTA Results with Long Training Time}
In PSGTR, every triplet is expected to be captured by a query, which needs to predict everything in the triplet simultaneously, so the model is required to better focus on the connections between objects.
Besides, the cross-attention mechanism in the transformer encoder and triplet decoder enable each triplet query access to the information of the entire image.
Therefore, PSGTR is considered as the most straightforward and simplest one-stage PSG baseline with minimal constraints or prior knowledge.
As a result, although PSGTR only achieves one-digit recall scores in 12 epochs, it surprisingly achieves SOTA results with a prolonged training time of 60 epochs.

\section{Challenges and Outlook}

\noindent\textbf{Challenges}\quad
While some \emph{prior knowledge} introduced by some two-stage SGG methods might not be effective in the PSG task, we expect that more creative knowledge-aided models can be developed for the PSG task in the era of multi-modality, so that more interesting triplets can be extracted with extra priors.
However, it should be noted that although priors can be useful to enhance performance, the PSG prediction should heavily rely on \emph{visual clues}. For example, in Fig.~\ref{fig:exp_seg}, \texttt{person-walking on-pavement} should be identified if the model can perceive and understand the subtle visual differences between \texttt{walking} and \texttt{standing}.
Also, PSG models are expected to \emph{predict more meaningful and diverse relations}, such as rare relations like \texttt{feeding} and \texttt{kissing}, rather than only being content with statistically common or positional relations.

\noindent\textbf{Relation between PSG and Panoptic Segmentation}\quad
Fig.~\ref{fig:exp_seg} visualizes the panoptic segmentation results of PSG methods, where PSGTR only obtains a miserable PQ result even with good PSG performance. 
The reason is that triplet queries in PSGTR produce object groundings independently, so that one object might be referred and segmented by several triplets, and the deduplication or the re-identification (Re-ID) process is non-trivial. Although the performance of Re-ID does not affect PSG metrics, it might still be critical to form an accurate and logical scene understanding for real-world applications.

\noindent\textbf{Outlook}\quad
Apart from attracting more research on the learning of relations (either closed-set or open-set) and pushing the development of scene understanding, we also expect the PSG models to empower more exciting downstream tasks such as visual reasoning and segmentation-based scene graph-to-image generation.

%% file: tables/exp_cls.tex
\begin{table}[t]
\caption{\textbf{Comparison between all baselines and PSGFormer.} 
	Recall (R) and mean recall (mR) are reported.
	IMP~\cite{xu2017scene} (CVPR'17), MOTIFS~\cite{zellers2018neural} (CVPR'18), VCTree~\cite{tang2018vctree} (CVPR'19), and GPSNet~\cite{lin2020gps} (CVPR'20) all originate from the SGG task and are adapted for the PSG task. Different backbones of ResNet-50~\cite{resnet} and ResNet-101~\cite{resnet} are used. 
	Notice that predicate classification task is not applicable to one-stage PSG models, so the corresponding results are marked as `$-$'.
	Models are trained using 12 epochs by default. $^\dagger$ denotes that the model is trained using 60 epochs.
}
\label{tab:exp_results}
\centering
\setlength\tabcolsep{2.5pt}
\resizebox{\textwidth}{!}{
\begin{tabular}{@{}clccc@{\hskip 0.1in}ccc@{\hskip 0.1in}ccc@{}}
\toprule
\multirow{2}{*}{Backbone} & \multirow{2}{*}{Method}  & \multicolumn{3}{c}{Predicate Classification} & \multicolumn{3}{c}{Scene Graph Generation}    \\ \cmidrule(l){3-8}
& & R/mR@20  & R/mR@50 & R/mR@100  & R/mR@20  & R/mR@50  & R/mR@100 \\ \midrule
\multirow{9}{*}{ResNet-50}
& IMP       & 31.9 / 9.55	& 36.8 / 10.9   & 38.9 / 11.6  & 16.5 / 6.52  & 18.2 / 7.05 & 18.6 / 7.23 \\
& MOTIFS    & 44.9 / 20.2   & 50.4 / 22.1   & 52.4 / 22.9  & 20.0 / 9.10  & 21.7 / 9.57 & 22.0 / 9.69 \\
& VCTree    & \textbf{45.3} / \textbf{20.5}   & \textbf{50.8} / \textbf{22.6}   & \textbf{52.7} / \textbf{23.3}  & \textbf{20.6} / \textbf{9.70}  & \textbf{22.1} / \textbf{10.2} & \textbf{22.5} / \textbf{10.2} \\
& GPSNet    & 31.5 / 13.2   & 39.9 / 16.4   & 44.7 / 18.3  & 17.8 / 7.03  & 19.6 / 7.49 & 20.1 / 7.67 \\
\cmidrule{2-8}
& PSGTR              & -  & - & -                              & 3.82 / 1.29 & 4.16 / 1.54 & 4.27 / 1.57 \\
& PSGFormer          & -  & - & -                              & \textbf{16.8} / \textbf{14.5} & \textbf{19.2} / \textbf{17.4} & \textbf{20.2} / \textbf{18.7} \\
\cmidrule{2-8}
& PSGTR$^\dagger$              & -  & - & -                    & \textbf{28.4} / \textbf{16.6} & \textbf{34.4} / \textbf{20.8} & \textbf{36.3} / \textbf{22.1} \\
& PSGFormer$^\dagger$          & -  & - & -                    & 18.0 / 14.8 & 19.6 / 17.0 & 20.1 / 17.6 \\
\midrule
\multirow{9}{*}{ResNet-101}
& IMP       & 30.5 / 8.97 & 35.9 / 10.5 & 38.3 / 11.3 & 17.9 / 7.35 & 19.5 / 7.88 & 20.1 / 8.02 \\
& MOTIFS    & 45.1 / 19.9 & 50.5 / 21.5 & 52.5 / 22.2 & 20.9 / 9.60 & 22.5 / 10.1 & 23.1 / 10.3 \\
& VCTree    & \textbf{45.9} / \textbf{21.4} & \textbf{51.2} / \textbf{23.1} & \textbf{53.1} / \textbf{23.8} & \textbf{21.7} / \textbf{9.68} & \textbf{23.3} / \textbf{10.2} & \textbf{23.7} / \textbf{10.3} \\
& GPSNet    & 38.8 / 17.1 & 46.6 / 20.2 & 50.0 / 21.3 & 18.4 / 6.52 & 20.0 / 6.97 & 20.6 / 7.17 \\
\cmidrule{2-8}
& PSGTR              & -  & - & -                              & 3.47 / 1.18 & 3.88 / 1.56 & 4.00 / 1.64 \\
& PSGFormer          & -  & - & -                              & \textbf{18.0} / \textbf{14.2} & \textbf{20.1} / \textbf{18.3} & \textbf{21.0} / \textbf{19.8} \\
\cmidrule{2-8}
& PSGTR$^\dagger$             & -  & - & -                     & \textbf{28.2} / \textbf{15.4} & \textbf{32.1} / \textbf{20.3} & \textbf{35.3} / \textbf{21.5} \\
& PSGFormer$^\dagger$          & -  & - & -                    & 18.6 / 16.7 & 20.4 / 19.3 & 20.7 / 19.7 \\
\bottomrule
\end{tabular}
}
\end{table}

%% file: figures/exp_seg.tex
\begin{figure}[t!]
    \centering
    \includegraphics[width=\linewidth]{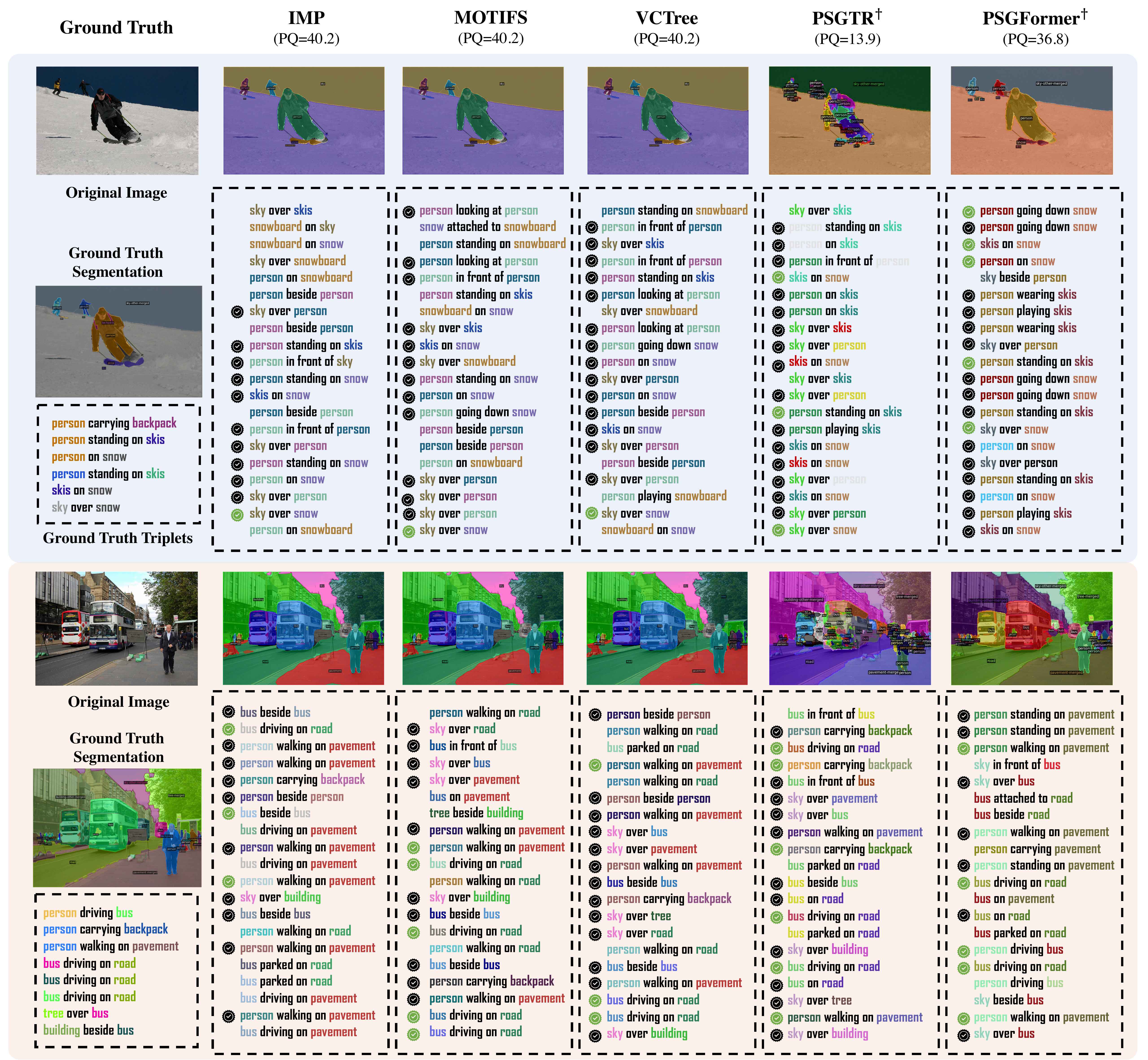}
    \caption{\textbf{Visualization of segmentations and the top 20 predicted triplets of 5 PSG methods.} The panoptic segmentation metric PQ is also reported.
    The colors of the subject and object in the triplet corresponds to the mask in the segmentation result.
    Reasonable triplets are marked by ticks.
    Triplets that match the ground-truth are marked by green ticks.
    One-stage models can provide more reasonable and diverse triplet predictions, but they are unable to achieve a good panoptic segmentation result.
    }
    \label{fig:exp_seg}
\end{figure}

%% file: sections/appendix.tex

\appendix
\setcounter{table}{0}
\renewcommand{\thetable}{A\arabic{table}}
\setcounter{figure}{0}
\renewcommand{\thefigure}{A\arabic{figure}}

\section{PSG Dataset Details}

\subsection{More comparisons between VG and PSG}
\begin{figure}[!h]
    \centering
    \includegraphics[width=0.9\linewidth]{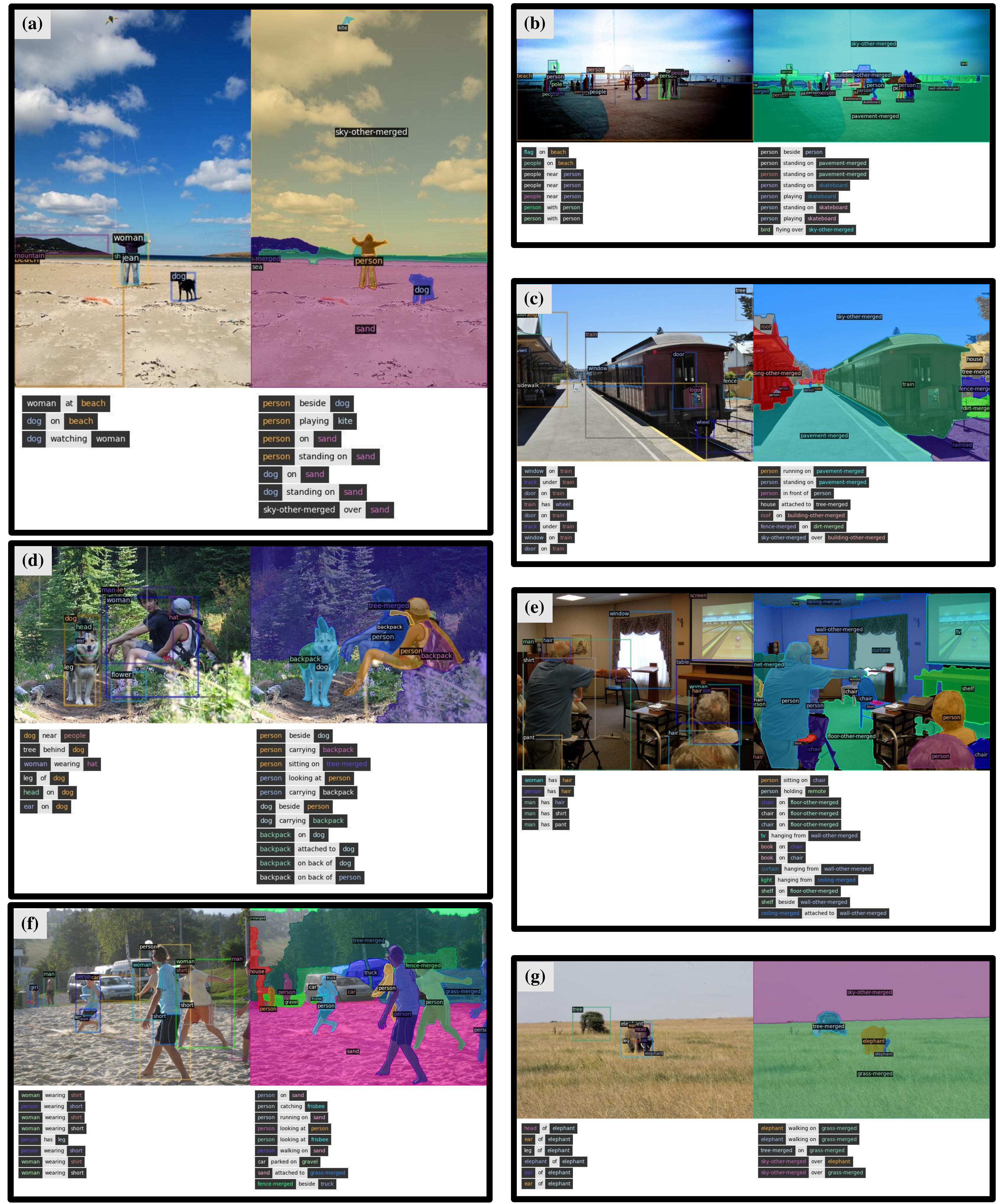}
    \caption{\textbf{More comparisons between VG-150 and PSG examples.} Every sub-figure has VG triplets and their groundings on the left and PSG on the right. Apart from precise pixel-wise grounding, PSG dataset gets rid of trivial relations (\eg, \texttt{head-on-dog} in (d), \texttt{person-has-hair} in (e)), and keep salient ones (\eg, \texttt{person-holding-remote} in (e)). Relations with background are also included (\eg, \texttt{elephant-walking-on-grass}).}
    \label{fig:more_example}
\end{figure}

More comparisons between VG-150 and PSG examples are shown in Fig.~\ref{fig:more_example}.
In particular, we would like to highlight some specific advances in the PSG dataset from sub-figure (a), which readers can confirm from other sub-figures.

In (a), VG-150 does not contain key information of `woman flying kite', and also has ambiguous relations like `at'. Fortunately, PSG addresses the key information but with a more general predicate `playing'. It is because in PSG, the predicate definition follows the rule of `being representative with proper granularity, not too specific'. Refer to Sec.~\ref{sec:supp_pred} for more information.
Also, PSG gathers far more comprehensive and accurate triplets.

For object groundings, it is noticeable that in VG, the grounding of `beach' is inaccurate, which only covers half of the actual beach. It can be problematic since a successful recall of a triplet requires a correct matching (big IOU) between predicted groundings and ground truth, in addition to a correct classification of triplets. Therefore, an incorrect annotation on object grounding can cause an inaccurate evaluation on scene graph generation too.
Apparently, object grounding of PSG is far more accurate than VG.

\subsection{PSG Dataset Statistics}
The PSG dataset has a total of 48,749 annotated images with 56 predicate classes, and 80 thing and 53 stuff classes (same as the COCO dataset~\cite{lin2014microsoft}).

Here is a list of average statistics for each image:
\begin{itemize}
    \item 11.0 instances per image
    \item 5.6 relations per image
    \item 1.9 (34\%) thing-thing relations per image
    \item 1.2 (21\%) stuff-stuff relations per image
    \item 2.5 (45\%) thing-stuff relations per image
\end{itemize}

\subsection{PSG Dataset Construction Details}

\paragraph{Built on COCO and Visual Genome}
In order to create a dataset with both panoptic segmentation \textit{and} relationship annotations, we took advantage of the overlap between the COCO~\cite{lin2014microsoft} and Visual Genome datasets~\cite{vg17ijcv}, where they share 48,749 images. Namely, for a given image, it has panoptic segmentation annotations from COCO, as well as scene graph annotations from Visual Genome. However, we cannot merge the two datasets directly as not only do they have different object annotations, they also define different object categories.
Therefore, we attempted to conduct the following dataset merging process.

\begin{figure*}[!t]
    \centering
    \includegraphics[width=0.8\linewidth]{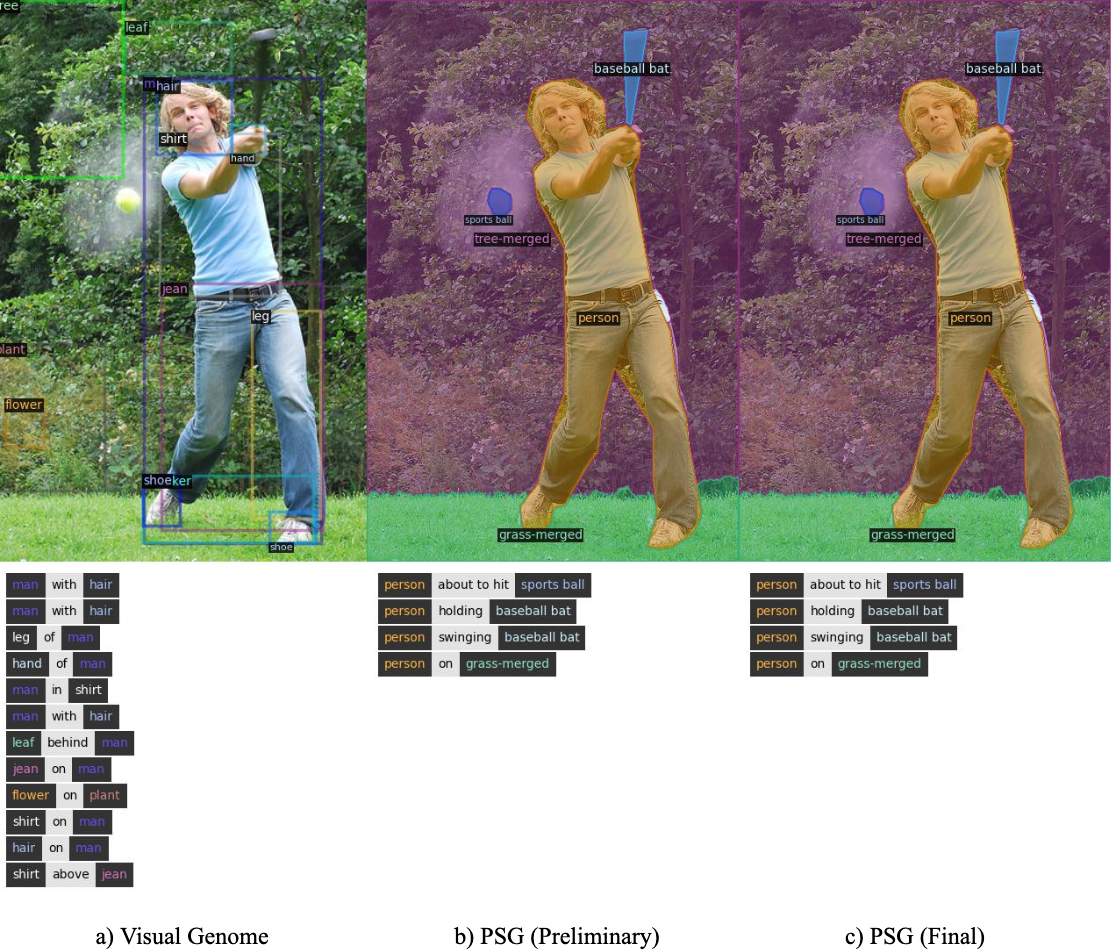}
    \caption{\textbf{PSG Dataset Constrution Process.} The object categories and annotations in (a) VG-150 are different from that in COCO (image in (b) shows the COCO panoptic segmentation). After matching COCO objects to VG objects using the object matching process, any corresponding relationship annotations can also be transferred over from (a) to (b). 
    These preliminary automatic but noisy relationship annotations are then sent to be processed by a final round of cleaning and annotation to produce the final PSG dataset in (c). } 
    \label{fig:eg_sg_annos}
\end{figure*}

\paragraph{Merging COCO and Visual Genome Annotations}
Dataset merging requires solving two intermediate tasks, saying 
1) \emph{Object Category Matching:} to figure out a mapping between the object categories in COCO to the object categories in Visual Genome, and 
2) \emph{Object Instance Matching:} for each image, to find out which object annotations in COCO correspond to which object annotations in Visual Genome. 

The goal of dataset merging is that, once the matching is achieved, we can transfer over the relationship annotations directly. 
However, since the matching process will not be completely perfect, we planned to bring these preliminary annotations to experienced and trained annotators for a final round of cleaning and annotation.

With a clear goal of dataset matching, we introduce the details of two intermediate tasks of object category matching and object instance matching. After the matching process, we can get a noisy preliminary PSG dataset (ref. Fig.\ref{fig:eg_sg_annos}(b)) that awaits the final cleaning.

\textbf{Object Category Matching:}
For a given COCO object category, what object categories in Visual Genome does it correspond to? For example, ``car'' in COCO may be matched to both ``car" and ``vehicle" in Visual Genome.

We encode the text of each object category into a feature vector using fastText~\cite{bojanowski2016enriching} word embeddings, and compute a similarity score for each (COCO, Visual Genome) object category pair using their cosine similarity. 

This similarity score will be useful for matching object instances in COCO to that in Visual Genome, as described in the next section.

\textbf{Object Instance Matching:}
The relationship annotations for each image in Visual Genome are tied to Visual Genome object annotations (object categories + bounding boxes), which are different from its corresponding COCO object annotations (object categories + panoptic segmentations). In order to transfer over the relationships to COCO object annotations, we can attempt to match each object instance as annotated in COCO, to an object instance as annotated in Visual Genome. Intuitively, if an object as annotated in COCO has a high overlap with an object as annotated Visual Genome, and their object categories are similar, they are likely to be referring to the same object. A sketch of the algorithm used to perform the matching is as follows.
For each image, we:
\begin{enumerate}
  \item compute the bounding box IoU of each object in COCO to each object in Visual Genome (using the tightest bounding box of the segmentation). 
  \item we then perform a greedy approach by always considering the instance pair with the highest IoU:
  \begin{enumerate}
      \item If their categories match, i.e. if the similarity score between the word embeddings of their category names are above a certain threshold, we'll \textbf{match} the pair together and remove them from the candidate pool.
      \item If the categories don't match, we \textbf{don't match} them and regard this pair as invalid.
      \item Move on to the next object pair with the highest IoU (start from 2. again). Repeat until there are no object pair candidates left, or if the remaining pairs have an IoU of 0.
  \end{enumerate}
\end{enumerate}

After the matching, the relationship annotations in Visual Genome can be transferred over to the COCO object annotations. This process is repeated for all the variants VG-150~\cite{xu2017scene}, GQA~\cite{hudson2019gqa} and VrR-VG~\cite{liang2019vrr}. This helps to maximize the recall of potentially correct scene graph relationships and alleviates the difficulty of the final annotation task for the annotators. 

\paragraph{Annotation Process}
Building upon the preliminary (noisy) PSG dataset (shown in Fig.~\ref{fig:eg_sg_annos}), we patiently trained our annotators to 
1) filter out incorrect triplets, and 
2) supplement more relations between not only object-object, but also object-background and background-background pairs, using the predefined 56 predicates.
The definition of 56 predicates will be explained in the next section.
The noisy triplets (for later filtering) are shown to be a good practice for annotators, prompting them providing both salient and detailed information.

\subsection{Predicate Dictionary}
\label{sec:supp_pred}
The design of predicate dictionary is inspired by COCO's practice on object categories selection. 
According to COCO, the selected categories must be representative, be relevant to practical applications, and be common with high occurrence.
Also, a proper level of granularity should also be considered.
With these principles in mind, we refer to all the predicates left in the preliminary PSG dataset, sorting them according to their occurrence, and carefully select the predicates. 
\paragraph{Practice for the Principles}
To meet the principles mentioned above, several processes are designed:
\begin{itemize}
    \item \textbf{For Representative:} we deduplicate the predicates and try to make the remaining predicates orthogonal. For example, we shrink a list of similar predicates of `parked along', `parked alongside', `parked at', `parked behind', `parked beside', `parked by', `parked in', `parked in front of', `parked near', `parked next to', `parked on' (existed in VG and GQA) to only keep one predicate as `\texttt{parked on}'. 
    Also, for the bidirected relation pairs such as `in front of' and 'behind', we only keep one direction, \ie, `\texttt{in-front-of}'. Similarly, only `\texttt{over}' is included while `beneath' is excluded.
    With this process, we make our vocabulary very concise and thus representative.
    \item \textbf{For Practicality:}
    Since the goal of the PSG dataset is to facilitate the development of scene understanding tasks, inspired by VrR-VG~\cite{liang2019vrr}, we get rid of many positional relations that fill the GQA dataset~\cite{hudson2019gqa}, such as `on the left of' and `on the right of', and especially focus on the visual-related predicates during our dictionary building.
    \item \textbf{For Coverage:} After several iterations, we finally decided to include 56 predicates with property and can well cover almost all the existing critical relations in the PSG dataset.
\end{itemize}

\paragraph{56 Predicates in the Dictionary}
\begin{itemize}
\item \textbf{Positional Relations (6):} 
over, in front of, beside, on, in, attached to.
\item \textbf{Common Object-Object Relations (5):} 
hanging from, on the back of, falling off, going down, painted on.
\item \textbf{Common Actions (31):} 
walking on, running on, crossing, standing on, lying on, sitting on, leaning on, flying over, jumping over, jumping from, wearing, holding, carrying, looking at, guiding, kissing, eating, drinking, feeding, biting, catching, picking (grabbing), playing with, chasing, climbing, cleaning (washing, brushing), playing, touching, pushing, pulling, opening.
\item \textbf{Human Actions (4):}
cooking, talking to, throwing (tossing), slicing.
\item \textbf{Actions in Traffic Scene (4):}
driving, riding, parked on, driving on.
\item \textbf{Actions in Sports Scene (3):}
about to hit, kicking, swinging.
\item \textbf{Interaction between Background (3):}
entering, exiting, enclosing (surrounding, warping in).
\end{itemize}

\paragraph{Detailed Predicate Definitions}
We provided a detailed explanation on each predicate with image examples to ensure the consistent performance from annotators. We will provide the handbook in our PSG website.

\begin{figure*}[!t]\RawFloats
\centering
\includegraphics[width=0.9\linewidth]{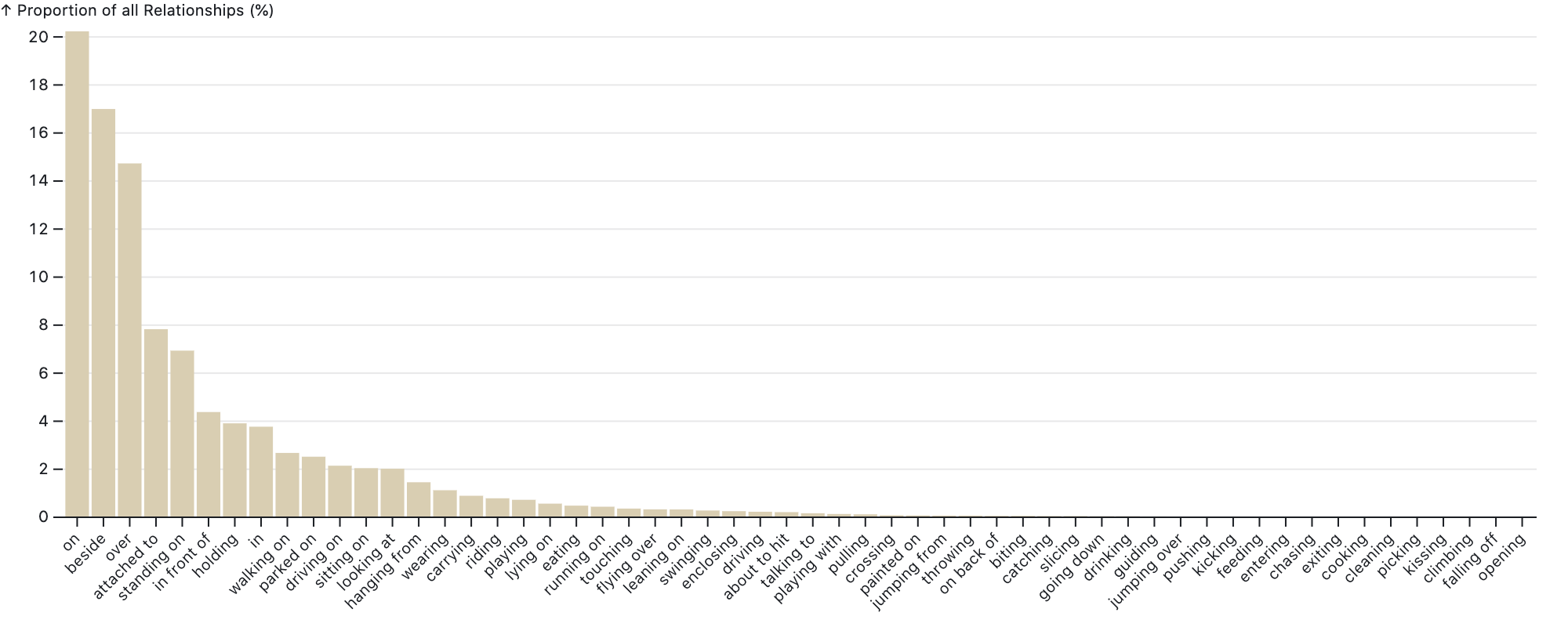}
\caption{\textbf{Proportion of Predicate Classes (Sum=100\%).}} 
\label{fig:eg_sg_annos}
\end{figure*}
\begin{figure*}[!t]\RawFloats
\centering
\begin{minipage}{0.43\textwidth}
\includegraphics[width=\linewidth]{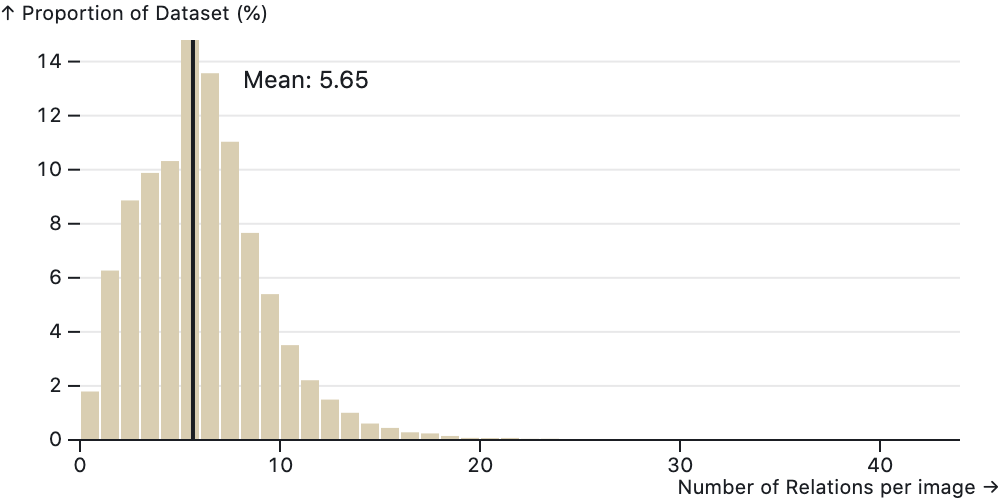}
\caption{\textbf{Distribution of the Number of Relations per image in the PSG Dataset.} The bulk of the images have around 5 - 10 relationship annotations, and ranges from 1 - 43 annotations. } 
\label{fig:eg_sg_annos}
\end{minipage}
\hspace{3mm}
\begin{minipage}{0.43\textwidth}
\includegraphics[width=\linewidth]{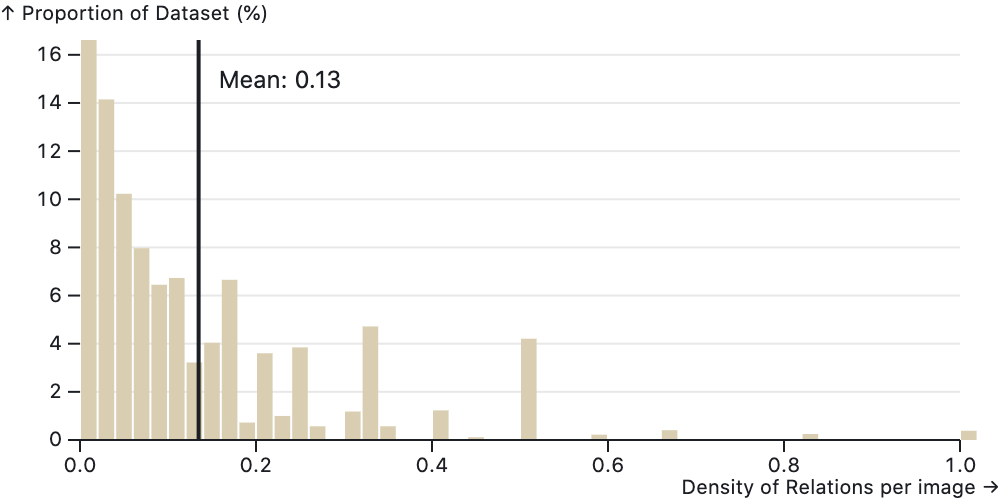}
\caption{\textbf{Distribution of the Density of Relations per image in the PSG Dataset.} The density of relations is defined as the number of annotated relations divided by the total number of possible relations in an image. } 
\label{fig:eg_sg_annos}
\end{minipage}
\end{figure*}

\section{Implementation Details}
All experiments are performed in a single unified codebase using the \texttt{MMDetection} framework to facilitate reproducibility.

\subsection{Two-Stage PSG Baseline}

\paragraph{Fine-tuning the Base Model}
We first fine-tune the Panoptic FPN base model on the panoptic segmentation annotations from our PSG dataset. The model is initialized from the best performing pretrained weights provided by \texttt{MMDetection}, and then trained using a batch size of 8. The SGD optimizer is used with a learning rate of 0.02, momentum of 0.9, weight decay of 0.0001, and gradient clipping with a max L2 norm of 35. Training runs for 12 epochs, with a learning rate schedule that linearly warms up from 0.02 / 3 to 0.02 over 500 iterations, and decays by a factor of 0.1 at the 8th and 11th epochs. 

\paragraph{Training the Scene Graph Prediction Head}
Using the fine-tuned Panoptic FPN above, we freeze its weights and only train the scene graph prediction head. This essentially treats the Panoptic FPN as a black-box feature extractor and panoptic segmentation predictor. For each predicted object, we extract a feature vector using RoIAlign (like in MaskRCNN), making use of the tightest bounding box around its segmentation mask. With grid features, and class predictions and bounding box localizations for each object at hand, we can feed these into any scene graph prediction head for training and prediction. We use a batch size of 16, and the SGD optimizer with a learning rate of 0.03, momentum of 0.9, and weight decay of 0.0001, and gradient clipping with a max L2 norm of 35. Training runs for 12 epochs, with a learning rate schedule that linearly warms up from 0.03 / 3 to 0.03 over 500 iterations, and decays by a factor of 0.1 at the 7th and 10th epochs. The hyperparameters for the MOTIFS, VCTree and GPSNet models all follow the same settings in their respective papers. 

\subsection{PSGTR}
As it is described in Section 4.2, our PSGTR model extends DETR to PSG task with new heads and a triplet Hungarian matcher. In detail, we implement each of those Feed Forward Networks (FFNs) by a 3-layer MLP, and each panoptic head, following DETR segmentation, consists of a multi-head attention layer and a 6-layer FPN-like CNN. Besides, the number of queries is set as 100 which indicates that 100 possible relations are predicted.

\paragraph{Training settings} 
In general, we follow most of the training strategies of DETR. We adopt the same AdamW optimizer with $10^{-4}$ learning rate and $10^{-4}$ weight decay for PSGTR except for the backbone which is trained with learning rate of $10^{-5}$. For initialization, we directly use COCO pretrained DETR to initialize the weights of our backbone and transformer.  Besides, we also generally follow DETR’s data augmentation which does cropping and resizing operations with settings such that the shortest side is at least 480 and at most 800 pixels while the longest at most 1333. However, it should be noted that when cropping images, we also filter the ground truth of bounding boxes and relations pairs that might be cropped. We train our model for 60 epochs with a step scheduler at epoch 40, and it finally takes us around 2 days to train on eight V100 GPUs with batch size 1.


\subsection{PSGFormer}
PSGFormer is built on the baseline of PSGTR so that most of the training details are shared.
In detail, PSGFormer also implements each FFN by a 3-layer MLP, and each panoptic head by a multi-head attention layer and a 6-layer FPN-like CNN as DETR does. 
Besides, the number of object queries and relation queries are set as 100. 
We follow the training hyperparameters of PSGTR including optimizer, learning rate, data augmentation, \etc.
Notice that PSGFormer also has an auxiliary task of pure panoptic segmentation with object decoder, the ratio between the main task on triplet supervision and the auxiliary panoptic segmentation supervision is 5 to 1.
We train our model for 60 epochs, taking around 2.5 days to train on eight V100 GPUs with batch size 1.


\section{Visualization of PSGTR Result Triplet-by-Triplet}
\begin{figure}[!t]
    \centering
    \includegraphics[width=\linewidth]{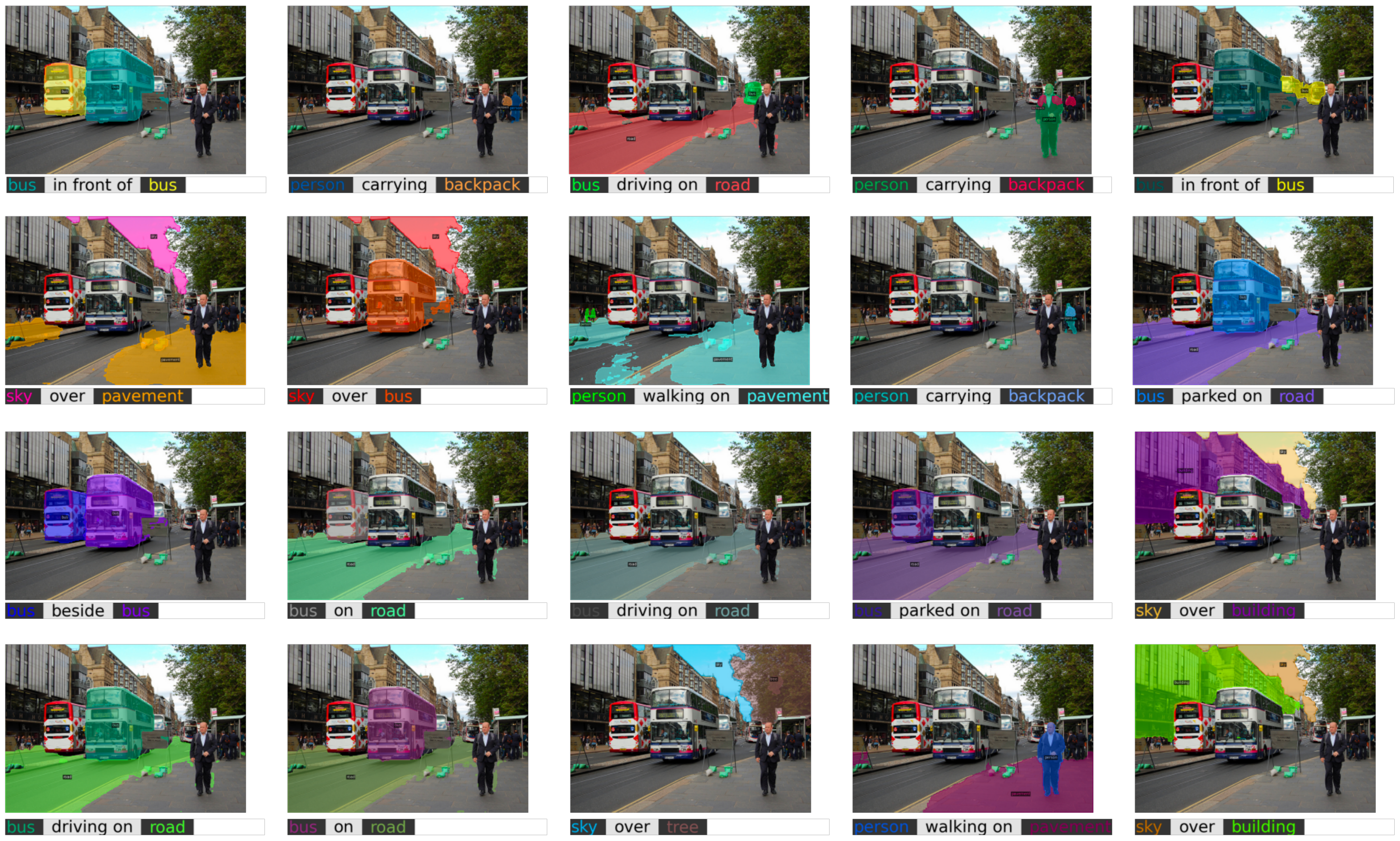}
    \caption{\textbf{Visualization of PSGTR Result Triplet-by-Triplet.} 
    PSGTR uses triplet queries to directly predict subject / object masks in the triplets, and the subject / object across triplets are not dependent, so the panoptic segmentation visualization is chaos in Fig.~\textcolor{linkcolor}{6}.
    However, if we visualize PSGTR result triplet-by-triplet, the result looks good.
    }
    \label{fig:psgtr_vis}
\end{figure}
Fig.~\ref{fig:psgtr_vis} shows PSGTR's predict results in a triplet-by-triplet fashion, as a complementary to the lower example in Fig.~\textcolor{linkcolor}{6}.
Notice that panoptic segmentation visualization is chaos in Fig.~\textcolor{linkcolor}{6}.
It is because PSGTR uses triplet queries to directly predict subject / object masks in the triplets, and the subject / object across triplets are not dependent, and the re-identification of each subject / object is no-trivial, and we use a simple post-processing method of pixel-wise argmax function to merge the segments, but it will still split one object into parts. However, it does not mean that PSGTR cannot segment objects well when predicting triplets. As we visualize the PSGTR result triplet-by-triplet in Fig.~\ref{fig:psgtr_vis}, the result looks good.